# Automatic Page Segmentation Without Decompressing the Run-Length Compressed Text Documents

Mohammed Javed[$] and P. Nagabhushan

Department of IT, Indian Institute of Information Technology, Allahabad, India-211015

[$]javed@iiita.ac.in

**Abstract**

Page segmentation is considered to be the crucial stage for the automatic analysis of documents with complex layouts. This has traditionally been carried out in uncompressed documents, although most of the documents in real life exist in a compressed form warranted by the requirement to make storage and transfer efficient. However, carrying out page segmentation directly in compressed documents without going through the stage of decompression is a challenging goal. This research paper proposes demonstrating the possibility of carrying out a page segmentation operation directly in the run-length data of the CCITT Group-3 compressed text document, which could be single- or multi-columned and might even have some text regions in the inverted text color mode. Therefore, before carrying out the segmentation of the text document into columns, each column into paragraphs, each paragraph into text lines, each line into words, and, finally, each word into characters, a pre-processing of the text document needs to be carried out. The pre-processing stage identifies the normal text regions and inverted text regions, and the inverted text regions are toggled to the normal mode. In the sequel to initiate column separation, a new strategy of incremental assimilation of white space runs in the vertical direction and the auto-estimation of certain related parameters is proposed. A procedure to realize column-segmentation employing these extracted parameters has been devised. Subsequently, what follows first is a two-level horizontal row separation process, which segments every column into paragraphs, and in turn, into text-lines. Then, there is a two-level vertical column separation process, which completes the separation into words and characters. The proposed method was experimented with and validated with a subset of 166 compressed versions of the text documents from the repository of the publicly available MARG dataset, reporting an overall page segmentation accuracy of 95% at the text line level.

## 1. Introduction

Automatic page segmentation in complex layout documents finds various applications in Document Image Analysis (DIA), such as OCRing [1], equivalence detection [2], reading order determination [3], and so on. The performance and accuracy of these applications are largely dependent on the efficiency of the page segmentation process. In the literature many page segmentation algorithms have been proposed. These algorithms can be broadly grouped into three categories; namely based on the analysis of document foreground regions, background regions, and both foreground and background regions [4]. Foreground-based methods perform page decomposition by analyzing the foreground pixels, which normally include printed contents [3]. Second, background-based methods group the background pixels (also called "white space") for page segmentation [1]. Lastly, the hybrid-based methods analyze both foreground and background pixels to carry out efficient page decomposition [4].

The page segmentation algorithms available in the literature generally work on uncompressed or decompressed documents. However, in reality, the documents in document processing systems, such as fax machines [5], Xerox machines, and digital libraries, are made available in a compressed form to provide better archival and communication efficiencies. This implies that the existing systems need to decompress the compressed document and then operate, which consumes additional computing resources. Therefore, it would be novel to think of developing methods that can directly operate over the compressed data to carry out the required operations without going through the stage of decompression. Recently, this novel idea of operating directly in a compressed domain has been endorsed by many researchers and has been proven to offer better time and space efficiencies in comparison with conventional methods [6-8]. Hence, the direct processing of data in a compressed domain has become a prime area of research.

The initial attempt of working directly over compressed data was made in the early 1980s by Grant and Reid [9] and Tsuiki et al. [10]. Run-Length Encoding (RLE), a simple compression method, was first used in coding pictures [11] and television signals [12]. Also, there are some efforts that have been seen for performing different operations in compressed representations, such as image rotation [13], connected component extraction [14], skew detection [15], page layout analysis [16], document similarity [17], equivalence detection [18], and retrieval [19]. One of the recent works using run-length data performs morphological related operations [20]. In most of these works, the researchers use either run-length information from the uncompressed image or do some partial decoding to perform the operations. A detailed study to show the direct operations in the compressed data of TIFF [21] documents from the viewpoint of computational costs, efficiency, and validation with a large dataset has not been attempted. However, there have been initiatives by [22,23] to show the feasibility of performing segmentation without decompression using JPEG documents, which may not directly operate over TIFF compressed text documents.

The research goal of this paper is to demonstrate the possibility of performing a page segmentation operation in compressed text documents. However, the work is limited to the compressed data of binary text documents. In the literature, there are different compression schemes available for compressing text document images. The popular compression schemes that are specifically designed for handling fax documents and handwritten text documents are CCITT Group-3 [24] and CCITT Group-4 [25], which are widely supported by TIFF and PDF image file formats. This paper is focused on how to employ the run-length compressed data of the CCITT Group-3 compression scheme to demonstrate automatic page segmentation. However, there is also a research plan to extend this page segmentation to CCITT Group-4 and other standard text document compression methods in the near future.

For many decades, page segmentation on uncompressed documents has remained an active area of research in document image analysis. The recent developments in this field can be obtained from [1,3,4]. In [3], based on the foreground region analysis, Zirari et al. have proposed a page decomposition method that employs the graph modeling of foreground pixel intensities to detect the connected components. Furthermore, they used the connected components to classify the document into text and non-text regions by applying structural layout rules. The researchers claimed that their method is superior in detecting the text portions within documents from the page segmentation dataset of UW-III and ICDAR-2009. On the other hand, based on the background region analysis in [1], Breuel has presented a column boundary detection algorithm for complex documents using whitespace rectangles. The whole idea is to detect the maximum empty rectangles within the document based on background structure analysis. He demonstrated the method on documents from the UW-III database. In [4], Chen et al. have presented a hybrid page decomposition algorithm that extracts the white space blocks between the connected components. Using the information from the foreground and background regions, the unwanted white space blocks are filtered, and the remaining white space blocks are grouped to form column separators using some predefined rules. Also, the ICDAR-2009 page segmentation dataset was used to prove the superiority of the proposed model. Nevertheless, in the case of compressed documents, Regentova et al. [16] have proposed an interesting work with compressed documents to achieve document layout analysis using JBIG compression. Knowledge about the JBIG encoding process was employed and horizontal smearing was performed, which was followed by connected component extraction. A hybrid strategy was used where a top-down analysis resulted in structural layout and a bottom-up analysis yielded connected components. Simple geometrical features extracted from the connected components were used to classify the document into text and non-text blocks, and the layout analysis was extended for detecting form type and form dropout. In order to demonstrate the proposed idea, Rengetova et al. [16] only used eight compressed documents from ITU [26] test images.

However, the real time applications that employ JBIG compression are limited, and the compression scheme has yet to find its place in the TIFF image format.

Overall, the research goal of this paper is to demonstrate the possibility of carrying out an automatic page segmentation operation in run-length compressed documents that may contain text regions with inverted text color. To achieve the said target, a novel method to detect the inverted text regions in the compressed document using the analysis of foreground and background pixels is proposed, and in the subsequent step, a toggling operation is performed to bring back the inverted text regions to normal mode. Furthermore, to identify single-/multi-column documents, the presence of any vertical column separator in the document is checked. If column separators are detected, then text column segmentation is performed. Finally, each column block is further decomposed into segments of paragraphs, each paragraph into text lines, each text line into words, and each word into characters. All of which are carried out in the compressed version of the document without going through the stage of decompression.

The rest of the paper is organized as follows: Section 2 discusses the problem background details. Section 3 elaborates on our proposed model for page segmentation. Section 4 reports on the experimental analysis with the proposed methods, and Section 5 concludes the paper with a brief summary.

## 2. Problem Background

### 2.1 Foreground-Background in a Document Image

The scanned documents from research articles, newspapers, magazines, etc., generally come with different foreground and background colors followed by a complex layout. This makes the entire process of page segmentation and the subsequent step of OCRing a challenging task [27]. Therefore, it is necessary to detect foreground and background regions before carrying out any document analysis. A detailed study regarding the detection and separation of foreground and background regions has been conducted by Nirmala and Nagabhushan [27]. However, in case of text documents, black text is generally printed over a white background, and the possible alternate representation could be white text over a black background, which is called the "inverted text color region" in this paper. Therefore, this research also involves the automatic detection and toggling of these inverted text color regions in the compressed document before the actual page segmentation. In CCITT Group-3 compression standard, the black and white color pixels are represented and encoded using "1" and "0," respectively, in the bitmap format. The terms that are related to these representations are defined as listed below.

**Definition 1. Normal text region** ($R_N$): A text document region in a standard form with black text printed over a white background. It can also be referred to as a "1/0 layout."

**Definition 2. Inverted text region** ($R_I$): A text document region that is complementary to the normal text region with white text printed over a black background. It can also be referred to as a "0/1 layout."

Generally, text documents have normal text regions. However, in order to highlight some important information in the document, the text contents are sometimes printed with inverted text color regions, which make automatic OCRing and image analysis difficult. Therefore, in this research, before carrying out page segmentation an algorithm for the automatic detection and recovery of any such inverted text color regions in the compressed document into a normal text region is proposed. The presence of these regions is detected by the analysis of foreground and background pixels using a 0-1 (white and black pixel) histogram, and the detected regions are corrected by a simple toggle operation, which is explained in the section below. The appearance of inverted text color regions at various locations is shown in Fig. 1. Furthermore, based on the span of the $R_I$ regions in the text document, the $R_I$ regions may be horizontally stretched across the page, as shown in Fig. 1(a), (b), and (d), or vertically stretched across the text columns of the document, as shown in Fig. 1(c). Finally, in our research, the assumption that the minimum span of the $R_l$ region cannot be less than the span of a text line in the document has also been made.

(a) Single column (b) Double column (c) Double column (d) Complex layout

**Fig. 1.** A document in various layouts showing different regions with change in the foreground and background color. (a) Single column, (b) double column, (c) double column, and (d) complex layout.

## 2.2 Compressed Document Data

The document compression scheme that we used is the CCITT Group-3, which is widely supported over TIFF and PDF file formats. A text document employing the CCITT Group-3 compression standard uses Modified Huffman (MH) coding, which internally uses RLE as a basic compression technique. Generally, a run is a sequence of similar value pixels and the number of these pixels represents its length. The representation of the image pixels in the form of a sequence of run values and its lengths is known as run-length encoding. In the case of binary documents, coding is done with an alternating sequence of run lengths of 0 (white) and 1 (black), which produces a compact binary code. Table 1 illustrates this compressed data, which is also known as a "compressed run-matrix." The compressed run-matrix consists of alternate columns of runs of 0's (white pixels) and 1's (black pixels), which are identified as odd columns (1,3,5,…) and even columns (2,4,6,…), respectively. A sample of an uncompressed and compressed document is shown in Fig. 2.

**Table 1.** Description of run-length compressed binary data

| Binary image data | Compressed data | | | | |
|---|---|---|---|---|---|
| | 1 | 2 | 3 | 4 | 5 |
| 00000000000000 | 14 | 0 | 0 | 0 | 0 |
| 00110000111110 | 2 | 2 | 4 | 5 | 1 |
| 01111000111110 | 1 | 4 | 3 | 5 | 1 |
| 01111000111110 | 1 | 4 | 3 | 5 | 1 |
| 01111000111110 | 1 | 4 | 3 | 5 | 1 |
| 00110000000000 | 2 | 2 | 10 | 0 | 0 |
| 10000000000000 | 0 | 1 | 13 | 0 | 0 |
| 10000000000000 | 0 | 1 | 13 | 0 | 0 |
| 00100001111100 | 2 | 1 | 4 | 5 | 2 |
| 01110001111100 | 1 | 3 | 3 | 5 | 2 |
| 01111001111100 | 1 | 4 | 2 | 5 | 2 |
| 01111100000000 | 1 | 5 | 8 | 0 | 0 |
| 00000000000000 | 14 | 0 | 0 | 0 | 0 |

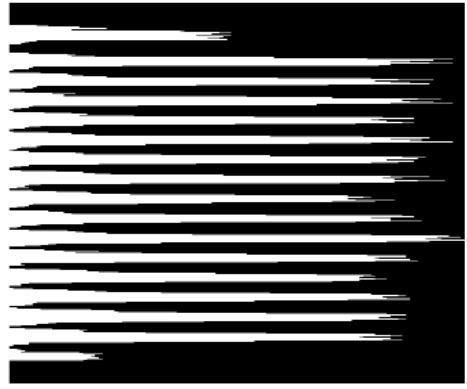

(a) Uncompressed Document    (b) Compressed Document

**Fig. 2.** A sample document data when viewed in uncompressed (a) and compressed (b) versions. The white content in (b) shows the presence of run in the compressed run-matrix, whereas the black portion indicates the sparsity within the run-matrix.

The compressed run-matrix in Table 1 consists of alternate columns of white pixel runs $W_{(i,j)}$ and black pixel runs $B_{(i,j)}$. Let the data structure of the compressed run matrix be denoted in a two-column format as $C(W_{(i,j)}, B_{(i,j)})$, where, $i = 1..m$ and $j = 1..n'/2$. Here, $m$ and $n'$ respectively denote the number of rows and columns present in the compressed document. $C(W_{(i,j)}, B_{(i,j)})$ for the compressed document provides access to two columns of the compressed run-matrix at a time. Therefore, the number of columns in the compressed document becomes $n'/2$.

In order to detect the inverted text regions in a compressed document, the row-wise computation of the histogram of foreground and background pixels, which is also known as a 0-1 histogram, is proposed. Generally, in a text document image for a given row scan of pixels, the number of background pixels is usually greater than the foreground pixels. Based on this notion, the histogram of black and white color pixels for each row is computed from the compressed data of the document. Let the number of black and white color pixels in the histograms be denoted as $H_B$ and $H_W$, respectively, and mathematically they can be computed from the compressed data as follows:

$$\text{Black pixel histogram } H_B(i) = \sum_{j=1}^{n'/2} B(i, j) \quad (1)$$

$$\text{White pixel histogram } H_W(i) = \sum_{j=1}^{n'/2} W(i, j) \quad (2)$$

**Table 2.** Foreground and background region detection with the analysis of black and white pixel histogram

| Analysis | Text region | Region indicator |
|---|---|---|
| $H_B > H_W$ | $R_I$ | $\delta = 1$ |
| $H_B < H_W$ | $R_N$ | $\delta = 0$ |

For a given row of runs in a compressed document, if $H_B$ is greater than $H_W$, it is confirmed that there are black pixels in the background and white pixels in the foreground. This type of row of runs represents the inverted text region, which is indicated by the notation $\delta = 1$. The rules formulated for detecting the $R_N$ and $R_I$ regions in the compressed data are summarized in Table 2. Before page segmentation, if any $R_I$ regions are detected, then these text regions have to be toggled to the standard $R_N$ text regions. This is achieved by row-wise shifting all of the runs towards one position left starting from the second column, which is expressed as follows:

$$R_N(i,j) = R_I(i,j+1), \text{ For all } j = 1..n'-1 \tag{3}$$

Another important issue in compressed data is generating vertical information. This is due to the fact that the run-length compressed data is horizontally compressed, which makes the tracing of vertical information a difficult task. Therefore, in order to extract the vertical information from the compressed data, we utilized a strategy proposed by Nagabhushan et al. [28]. Here, the vertical column information is generated virtually by popping out the run-length data in an intelligent sequence, as illustrated with an example of compressed data of the first six iterations, as shown in Table 3.

As seen in Table 3, the first two columns of the compressed data are used to generate vertical information. Let these two columns be represented by $C(W_{(i,1)}, B_{(i,1)})$ and the notation $\tau$ indicate the popped out transition value in each row. The different operations of Pop, Shift, and Terminate are defined with the help of these two column data structures in intelligently generating the vertical information from the horizontally compressed data. These operations are defined mathematically as:

$$Pop(W_{(i,1)}) \leftarrow C(W_{(i,1)}, B_{(i,1)}), \text{ if } W_{(i,1)} \neq 0, \text{ where } Pop(W_{(i,1)}) \leftarrow \begin{cases} \tau = 0, and \\ C(W(i,1), B(i,1)) = C(W(i,1), B(i,1)) - (1,0) \end{cases} \tag{4}$$

$$Pop(B_{(i,1)}) \leftarrow C(W_{(i,1)}, B_{(i,1)}), \text{ if } W_{(i,1)} = 0 \text{ and } B_{(i,1)} \neq 0, \text{ where } Pop(B_{(i,1)}) \leftarrow \begin{cases} \tau = 1, and \\ C(W(i,1), B(i,1)) = C(W(i,1), B(i,1)) - (0,1) \end{cases} \tag{5}$$

$$\text{Shift } C(W_{(i,1)}, B_{(i,1)}) \leftarrow \begin{cases} C(W(i,j), B(i,j)) = C(W(i,j+1), B(i,j+1)), \forall i = 1..\frac{n'}{2}-1 \\ If, W(i,1) = 0 \& B(i,1) = 0 \end{cases} \tag{6}$$

$$\text{Terminate } C(W_{(i,1)}, B_{(i,1)}) \leftarrow \begin{cases} If, C(W(i,1), B(i,1)) = (0,0) \& C(W(i,2), B(i,2)) = \phi \\ \forall i = 1..m \end{cases} \tag{7}$$

## 3. Proposed Model

The proposed model for automatic page segmentation in the run-length compressed data of text documents containing inverted text regions is shown in Fig. 3. The run-length compressed data is fed as input to the algorithm. The different stages involved in the page decomposition process are as explained below.

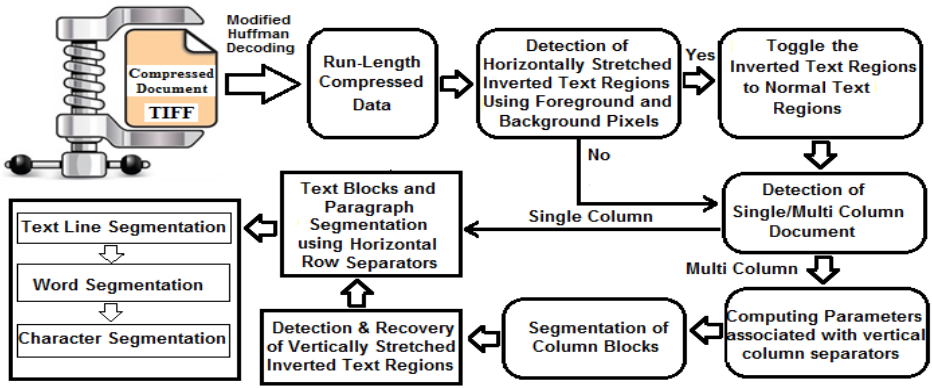

**Fig. 3.** Proposed model for page segmentation in compressed documents containing inverted text.

**Table 3.** Virtually generating column details from horizontally run-length compressed data [28]

| Pass | Line | Popped transition ($\tau$) | $W_{i,1}$ 1 | $B_{i,1}$ 2 | 3 | 4 | 5 | Status |
|---|---|---|---|---|---|---|---|---|
| Start | 1: |   | 14 | 0 | 0 | 0 | 0 |   |
|   | 2: |   | 2 | 2 | 4 | 5 | 1 |   |
|   | 3: |   | 1 | 4 | 3 | 5 | 1 |   |
|   | 4: |   | 1 | 4 | 3 | 5 | 1 |   |
| 1 | 1: | 0 | 13 | 0 | 0 | 0 | 0 | Pop |
|   | 2: | 0 | 1 | 2 | 4 | 5 | 1 | Pop |
|   | 3: | 0 | 0 | 4 | 3 | 5 | 1 | Pop |
|   | 4: | 0 | 0 | 4 | 3 | 5 | 1 | Pop |
| 2 | 1: | 0 | 12 | 0 | 0 | 0 | 0 | Pop |
|   | 2: | 0 | 0 | 2 | 4 | 5 | 1 | Pop |
|   | 3: | 1 | 0 | 3 | 3 | 5 | 1 | Pop |
|   | 4: | 1 | 0 | 3 | 3 | 5 | 1 | Pop |
| 3 | 1: | 0 | 11 | 0 | 0 | 0 | 0 | Pop |
|   | 2: | 1 | 0 | 1 | 4 | 5 | 1 | Pop |
|   | 3: | 1 | 0 | 2 | 3 | 5 | 1 | Pop |
|   | 4: | 1 | 0 | 2 | 3 | 5 | 1 | Pop |
| 4 | 1: | 0 | 10 | 0 | 0 | 0 | 0 | Pop |
|   | 2: | 1 | 0 | 0 | 4 | 5 | 1 | Pop |
|   | 3: | 1 | 0 | 1 | 3 | 5 | 1 | Pop |
|   | 4: | 1 | 0 | 1 | 3 | 5 | 1 | Pop |
| 5 | 1: | 0 | 9 | 0 | 0 | 0 | 0 | Pop |
|   | 2: | 0 | 3 | 5 | 1 | 0 | 0 | Shift-pop |
|   | 3: | 1 | 0 | 0 | 3 | 5 | 1 | Pop |
|   | 4: | 1 | 0 | 0 | 3 | 5 | 1 | Pop |
| 6 | 1: | 0 | 8 | 0 | 0 | 0 | 0 | Pop |
|   | 2: | 0 | 2 | 5 | 1 | 0 | 0 | Pop |
|   | 3: | 0 | 2 | 5 | 1 | 0 | 0 | Shift-pop |
|   | 4: | 0 | 2 | 5 | 1 | 0 | 0 | Shift-pop |

### 3.1 Detection of Inverted Text Regions

The first stage before the page segmentation process is to detect the presence of inverted text regions by analyzing the foreground and background pixels by computing a 0-1 pixel histogram. Intuitively, the text documents generally have a larger number of background pixels than the foreground ones, except at the base point of a text line. This property is exploited and the 0-1 pixel histogram is computed for every row of the compressed data. Based on the rules formulated in Table 2, every row of the compressed document is marked with the region indicator $\delta$. The consecutive rows marked with $\delta = 1$ indicate the inverted text regions $R_I$. However, any abrupt change in the $\delta$ value, as observed at the base of a text line, is resolved using the $\delta$ value of its neighboring rows. The 0-1 pixel histograms for a document with inverted text regions in Fig. 1(b) are shown in Fig. 4, and the 0-1 pixel histograms for a document without inverted text regions are shown in Fig. 5.

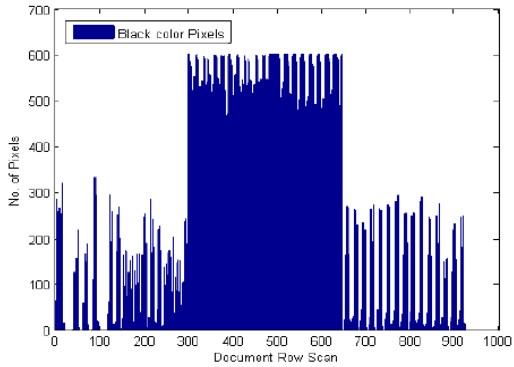
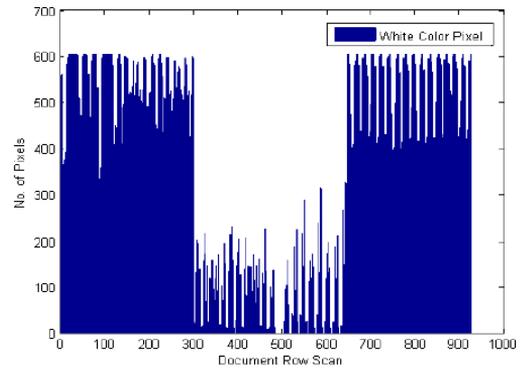

(a) Histogram of Black pixels  (b) Histogram of White pixels

**Fig. 4.** Histograms computed for black (a) and white (b) color pixels for every row in the compressed data of the document in Fig. 1(b).

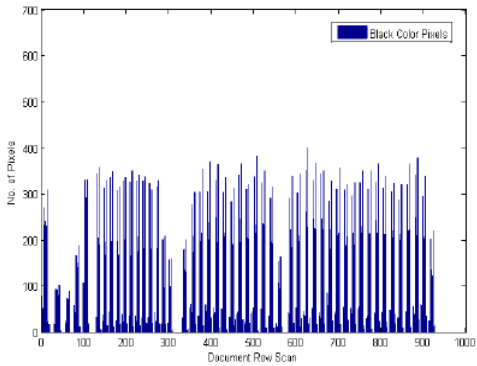
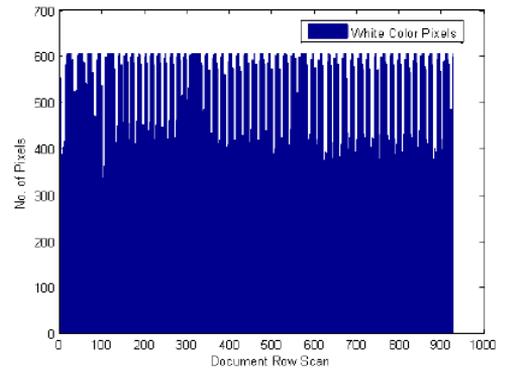

(a) Histogram of Black pixels  (b) Histogram of White pixels

**Fig. 5.** Histograms computed for black (a) and white (b) color pixels for every row in compressed data of document without inverted text region.

### 3.2 Toggling the Inverted Text Regions to the Standard Form

The presence of inverted text regions within a compressed document increases the complexity of automatic OCRing and document image analysis. Therefore, toggling the detected inverted text regions ($R_I$) that were identified in the previous stage to the normal text regions ($R_N$) is proposed. In the compressed data, toggling from $R_I$ to $R_N$ can be achieved by just shifting all the runs in the row to one column or one position to the left, as described in Section 2. A sample of a compressed document with different inverted text regions and its toggled version, which has been decompressed, is shown in Fig. 6.

**Fig. 6.** A sample document with different $R_I$ and its equivalent in $R_N$ after toggle operation. (a) Document with $R_I$ and (b) document converted to $R_N$.

### 3.3 Detection of a Single-/Multi-Column Document

After the toggling of horizontally stretched inverted text regions to normal text regions, the document is subjected to the detection of the presence of vertical column separators. A vertical column separator is a contiguous sequence of white runs of sufficient length and width in the vertical direction. A compressed document is said to be of single column if it does not contain any vertical column separators, and in the presence of one or more vertical column separators, the document is said to be multi-column. Therefore, the detection of a single-/multi-column document is achieved by detecting the number of vertical column separators in the document.

### 3.3.1 Detection of vertical column separators

A compressed document data that does not have any horizontally stretched $R_I$ regions is used as input for detecting the vertical column separators. Due to horizontal compression in run-length data, the vertical information becomes difficult to trace and the problem of detecting vertical column separators becomes challenging. Therefore, in order to generate the vertical information from the horizontally compressed data, a method [28] of generating the vertical columns virtually by popping out the horizontal runs in an intelligent sequence has been utilized, which has been discussed and illustrated in Section 2.

Once the entire vertical column is popped out from the horizontally compressed data in a sequence, the transition values within the column are compressed and checked for the presence of any white pixel run that is greater than the length threshold $L^v_{cs}$ of the vertical column separator. This threshold has been experimentally defined, as discussed later in Section 4. The popped-out columns that fulfill the aforementioned criteria are retained and the others are discarded. This process generates a sequence of contiguous columns in place of vertical column separators in the document and they are identified as candidates that are likely to be column separators. However, in a text document, there are chances to get contiguous columns of a very small width. Therefore, a threshold for the width of column separators, called $W^v_{cs}$, has also been defined experimentally and the likely to be column separators are filtered based on this threshold. Different possible column separators detected at this stage are shown in sample documents in Fig. 7. Also, any such column separator detected at the beginning and/or at the end of the document is removed with an assumption that the column separators cannot exist at these positions. The entire process results in detecting the vertical column separators from the compressed document.

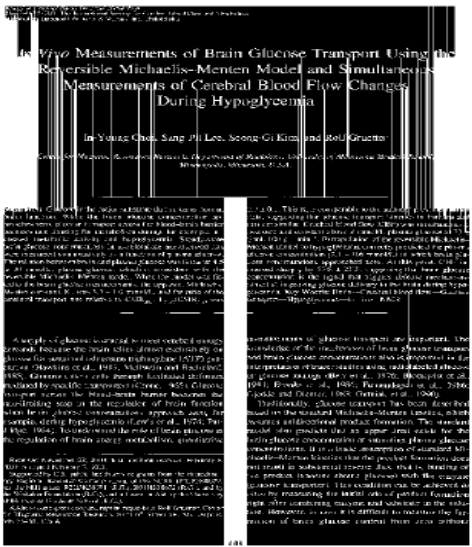
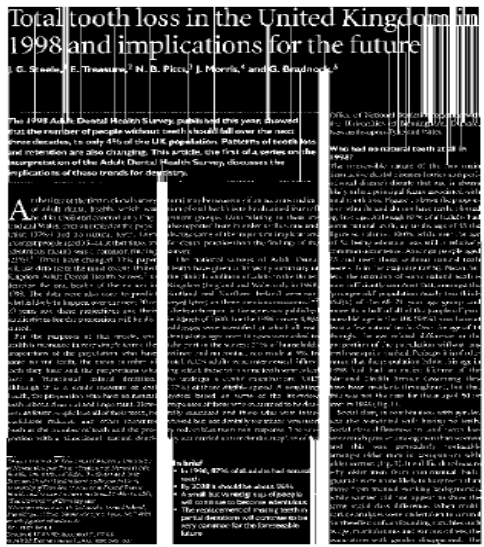

(a) Document with single columns separators    (b) Document with two columns separators

**Fig. 7.** Vertical column separators detected in sample documents (shown in decompressed version and the document images are printed with inverse color for the sake of clear visibility of vertical runs). (a) Document with single column separators and (b) document with two columns separators.

### 3.3.2 Computation of related parameters from the vertical column separator

After the detection of vertical column separators, the related parameters associated with them have to be computed from its compressed data, which will be used during the text column segmentation process. The related parameters of a vertical column separator are its length ($L_v$), horizontal start ($S_h$), and end ($E_h$) positions and the vertical begin ($V_\alpha$), and end ($V_\beta$) positions, which are illustrated in an uncompressed version of a sample document in Fig. 8.

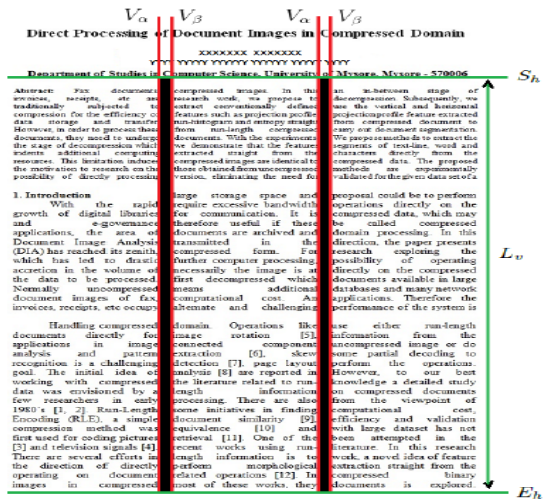

**Fig. 8.** Vertical column separators and relative parameters in a compressed version of a sample document.

Let the vertical compressed data of a column separator be denoted as $C'(x,y)$, where $x = 1..m'$ rows, and $y = V_\alpha..V_\beta$, which represents the begin and end column of the vertical column separator. In the compressed data of $C'(x,y)$, let $p$ represent the position of the longest run that contributes as the column separator. Therefore, the average length of the vertical column separator $L_v$ is computed by taking the average length of runs that contribute as the vertical column separator. This is computed as follows:

$$L_v = \frac{\sum_{y=V_\alpha}^{V_\beta} C'(p, y)}{V_\beta - V_\alpha + 1} \tag{8}$$

On the other hand, the horizontal start position $S_h$ of the vertical column separator is computed by taking the average of all the runs present above the run $p$ of the column separator. Whereas, the horizontal end location $E_h$ of the vertical column separator is the sum of its horizontal start position $S_h$ and vertical length $L_v$. The horizontal positions $S_h$ and $E_h$ can be mathematically represented as follows:

$$S_h = \frac{\sum_{y=V_\alpha}^{V_\beta} \sum_{x=1}^{p-1} C'(x, y)}{V_\beta - V_\alpha + 1} \tag{9}$$

$$E_h = S_h + L_v \tag{10}$$

## 3.4 Column Block Segmentation

The related parameters computed from the vertical column separators provide the row ($S_h$ and $E_h$) and column ($V_\alpha$ and $V_\beta$) details of the adjacent text column blocks. With the knowledge about the dimensions of each text column block, the compressed data of the relative column blocks is extracted using the block segmentation algorithm proposed by the authors in [29] for run-length compressed data. After the successful segmentation of the text column blocks, there are chances that the segmented text column block may contain vertically stretched $R_I$ regions, as shown in Fig. 1(c), which goes undetected in the first stage due to its small span when compared to the width of the document. Therefore, in the subsequent step, the segmented text column block is again subjected to the detection and recovery of $R_I$ regions using the same principle of 0-1 histogram analysis, as discussed earlier. At this stage, any $R_I$ region that is stretched across the text column block will be detected and toggled to the $R_N$ region. Finally, the inverted text region frees the compressed data in the document and this data is subjected to text block and paragraph segmentation, which is discussed in the next section.

## 3.5 Text Block Detection and Segmentation

After segmenting all of the text column blocks from the compressed document, each column block is further decomposed into text blocks and paragraphs. A text block is a block of text lines separated with horizontal text line separators of equal width $W_{ls}$. On the other hand, a horizontal text block separator of width $W_{bs}$ separates each text block. In the compressed data, a text block separator is a contiguous group of horizontal white space runs having a run-length greater than or equal to the length of the text line of that column block. A text block may contain one or more paragraphs that are either separated by horizontal paragraph separators of width $W_{ps}$ and/or by an indentation $I_p$. The different horizontal row separators in a compressed text document are shown with an uncompressed version of a column block in Fig. 9. The process of segmenting the entire text column block into paragraphs and text lines with horizontal row separators is known as coarse level segmentation; whereas, segmenting a text line into word and character segments with vertical column separators is known as fine level segmentation. Text line and fine level segmentations are discussed in the next section.

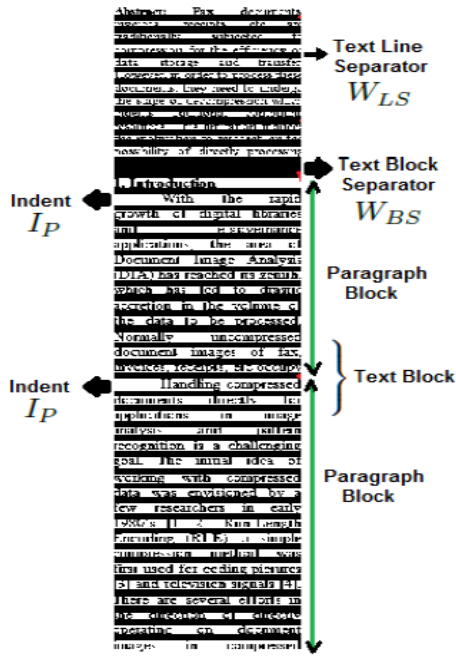

**Fig. 9.** Detection of text blocks and paragraphs in a segmented column block. The black strips indicate the spacing between text lines, and based on its width the text block separators are detected. Further, the presence of indent $I_p$ within the identified text blocks is used to locate paragraphs.

### 3.6 Text Line, Word, and Character Segmentation

In the final stage, the compressed data of every paragraph is further broken down into compressed segments of text lines, words, and characters. Every text line is separated with a text line separator of width $W_{ls}$ by which text line segmentation is achieved. The text line separators are detected by employing the popular vertical projection profile technique for compressed documents [8,30].

Every word in a text line is separated with a vertical column separator of width $W_{ws}$ and, similarly, every character has vertical column separator of width $W_{cs}$. To carry out word and character segmentation, the same analogy of vertical column separator detection proposed in this research work has been utilized. However, the detection of the vertical column separator was carried out for a single text line. We borrowed the same idea by the authors in [30] to achieve word and character segmentation from the compressed text block based on the width threshold of vertical column separators for the words $W_{ws}$ and characters $W_{cs}$ of a text line.

### 3.7 Other Related Special Issues

The stages discussed above summarize the overall general procedure for the proposed page segmentation method. However, there are some special cases that need to be tackled during the process of detecting column separators in the compressed document data. Consider the case of detecting column separators in a document with a complex layout that is similar to the document shown in Fig. 1(d). For the sake of demonstration, the different column separators detected for this sample document have been marked within a rectangular strip in the uncompressed version of the document in Fig. 10. Although there are six vertical column separators, the only five column separators are detected by the proposed algorithm. This is because of the presence of two overlapping column separators within a single contiguous group of columns. In this case, the proposed method is extended to detect the presence of overlapping column separators and column block segmentation is performed. Initially, all of the possible column separators present in the document are detected. During the detection process, if two or more larger runs in a column satisfy the specified threshold parameters of the vertical column separator, then this group of columns is identified as being special overlapping column separators.

Once all of the unique and special overlapping column separators are identified, column block segmentation is carried out. For a complex layout document, like the one shown in Fig. 1(d), a strategy is employed in order to extract the column blocks in an ordered sequence. The first column separator is used as a reference point in the first iteration and the mid-location of the other remaining column separators is computed with the formula:

$$\text{Mid point } M_h = \frac{E_h - S_h + 1}{2} \quad (11)$$

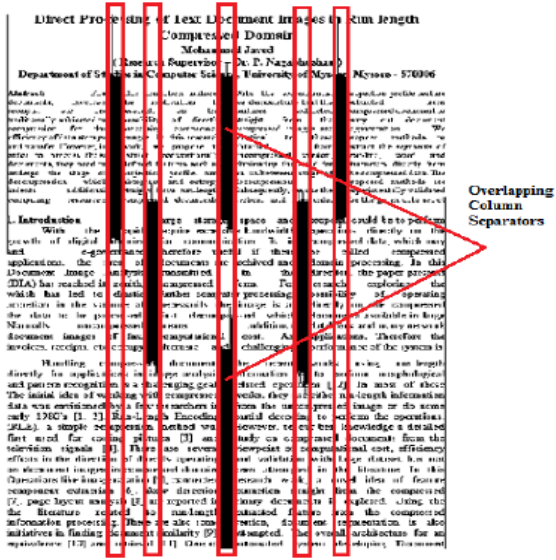

**Fig. 10.** Detection of vertical column separators in document with complex layout.

All of the column blocks that have their column separator midpoint between the horizontal start ($S_h$) and end point ($E_h$) of the reference column separator are segmented, and they form a single set of horizontally aligned column blocks. In the second iteration, out of the remaining column separators, the first column separator is made as the reference column separator and the aforesaid procedure is repeated to segment the next sequence of column blocks. Meanwhile, it should be noted that a pointer is maintained for special overlapping column separators to keep track of segmented blocks from the previous iterations. This process is repeated until all of the column separators are used.

## 4. Experimental Results and Discussions

In this section, the discussions regarding the dataset that we used for experiments, the different performance measures that we used for evaluating the proposed algorithms, and the different learning and testing parameters that we applied are presented.

### 4.1 Dataset Specifications and Performance Evaluation

For our experiments, we chose a subset of 104 uncompressed text documents of single- and multi-column layouts from the publicly available Medical Article Records Ground (MARG) truth dataset [31], which we denoised, deskewed, and compressed. The MARG dataset consists of 1,553 uncompressed documents obtained from the first pages of medical journals and where the majority of the documents are in single column and double column layouts. The subset of documents that we selected from various journals covers most of the variations that may occur within a single and double column layout of a text document. Furthermore, six documents containing inverted text color regions were collected from the MARG dataset itself. However, to check the performance of the proposed algorithm, 56 randomly selected documents from the selected subset of documents were modified to add some inverted text color regions. Altogether, we used a dataset of 166 compressed text documents for our experimental study.

In order to evaluate the performance of the proposed algorithm, the accuracy of the segmented text lines was used as a performance metric. The accuracy of detecting and toggling $R_I$ regions was measured in terms of the number of text lines correctly detected in the $R_I$ regions to the total number of text lines actually present in the $R_I$ regions, which was taken to be the number of ground truth text lines. This is expressed as:

$$A_{it} = \frac{R_{gt} - R_{er}}{R_{gt}} \tag{12}$$

where, $R_{gt}$ indicates the number of text lines taken as ground truth text lines in $R_I$ regions, and $R_{er}$ is the number of errors or text lines not detected in the $R_I$ regions.

Similarly, on the other hand, in order to evaluate the performance of the overall page segmentation algorithm that includes inverted text color regions, the performance metric of text-line accuracy, which was proposed by [32], was used. It was redefined as page segmentation accuracy for compressed documents and is represented as:

$$A_{ps} = \frac{L_{gt} - L_{er}}{L_{gt}} \tag{13}$$

where, $L_{gt}$ is the total number of ground truth text lines and $L_{er}$ is the total amount of text lines that were wrongly detected.

## 4.2 Parameter Learning, Results, and Error Analysis

Out of the 104 compressed text documents selected for validating our proposed method, the first 50 documents were single column, 50 were double column, and the remaining were three columns. For training the algorithm, 10 documents from a single column layout and 10 from the double column layout were chosen. The different page segmentation parameters that were learnt experimentally from the training set are as follows:

1. Vertical column separator length threshold $L^v_{cs}$: {$m/6$ pixels, where $m$ is the number of rows in the compressed Document}
2. Vertical column separator width threshold $W^v_{cs}$: {70–120 pixels}
3. Horizontal text block separator $W_{bs}$: {$\geq 25$ pixels}
4. Horizontal paragraph separator $W_{ps}$: {10–25 pixels}
5. Horizontal text line separator $W_{ls}$: {$< 10$ pixels}
6. Paragraph Indent $I_p$: {30–100 pixels}
7. Word separator width threshold [30] $W_{ws}$: {$\geq 5$ pixels}
8. Character separator width threshold [30] $W_{cs}$: {$< 5$ pixels}

The experimental results obtained for the detection and toggling of $R_I$ regions with the dataset are shown in Table 4. In order to check the reliability of the algorithm, we experimented with our proposed method again on compressed documents containing non-text components, and the results obtained are shown in Table 5. As per the experiment, the presence of non-text components, such as figures, tables, and graphs, did not affect the detection of inverted text color regions within the compressed document. However, figures that contained black pixel dense regions, which look similar to $R_N$ regions, were converted to $R_I$ regions, as shown in Fig. 11(a). The sample results obtained for detecting and correcting the $R_I$ regions in compressed documents that contain non-text components are shown in a decompressed version in Fig. 11.

**Table 4.** Accuracy of detection and correction of inverted text color regions in pure text documents

| Total documents | $R_{gt}$ | $R_{er}$ | $A_{it}$ (%) |
|---|---|---|---|
| 62 | 463 | 00 | 100 |

**Table 5.** Accuracy of detection and correction of inverted text color regions in documents containing non-text (images, tables and graphs)

| Total documents | $R_{gt}$ | $R_{er}$ | $A_{it}$ (%) |
|---|---|---|---|
| 10 | 101 | 00 | 100 |

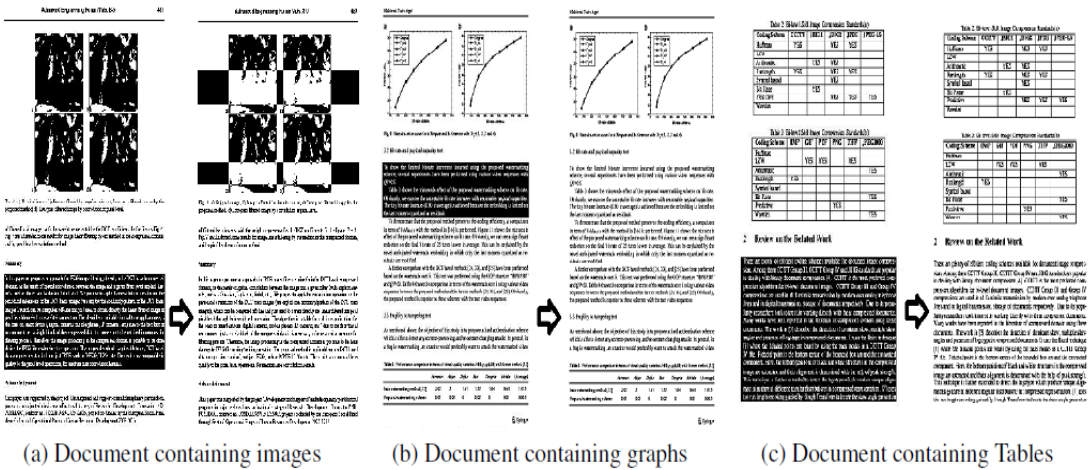

(a) Document containing images  (b) Document containing graphs  (c) Document containing Tables

**Fig. 11.** Results obtained for detection and correction of $R_I$ regions in compressed text documents in the presence of non-text components (shown in decompressed version).

**Table 6.** Accuracy of page segmentation

| Dataset | Total documents | $L_{gt}$ | $L_{er}$ | $A_{ps}$ (%) |
|---|---|---|---|---|
| Test documents | 146 | 9772 | 426 | 95.64 |
| Test+Training documents | 166 | 10950 | 501 | 95.42 |

The accuracy of the overall page segmentation algorithm with the 166 document dataset is shown in Table 6.

Overall, the proposed page segmentation algorithm with compressed documents resulted in an approximate accuracy of 95% and produced 5% errors with the modified MARG dataset. From the experiments that we conducted, the following observations were made. The accuracy of page segmentation algorithm largely depends on the length threshold $L^v_{cs}$ of the vertical column separator and the parameters computed from the vertical column separators. If the value of $L^v_{cs}$ is increased, then the column separators of a small length will be excluded. Conversely, if the $L^v_{cs}$ threshold is decreased, it will produce false vertical column separators within the text document, as shown in Fig. 7. Therefore, the threshold parameter is largely dependent on the type of document being considered for segmentation. However, the length parameter computed from the detected vertical column separator also affects the accuracy of page segmentation. As observed in Fig. 7, the run-length of the white space (background color) that constitutes the vertical column separator is usually greater than the height of the text columns to be segmented. Therefore, the average length of vertical column separator computed from these runs sometimes results in an oversegmentation of the text columns. The performance of existing state-of-the-art algorithms for page segmentation for uncompressed text documents without any inverted text regions was evaluated by Mao and Kanungo [32]. They reported on the accuracy of the algorithms that they evaluated as being in the range 84%–95%, based on the same performance metric used in this paper for evaluating our own proposed algorithms.

**Fig. 12.** Sample document taken to study the performance of the proposed page segmentation algorithm (dimension 3081×2201 and three column layout).

We compared the performance of our proposed page segmentation algorithm with the conventional approach (decompression+segmentation) by using sample document, which is shown in Fig. 12. The performance metrics used were accuracy, time, and space, and the results obtained are shown in Table 7.

**Table 7.** Performance analysis of the proposed page segmentation algorithm for the compressed version of the sample document shown in Fig. 12

|  | Conventional approach (w.r.t. ground truth) | Proposed approach (w.r.t. ground truth) | Proposed approach (w.r.t. conventional) |
|---|---|---|---|
| Method | Decompression + Segmentation | Data analysis + Segmentation |  |
| Accuracy (%) | 96.33 | 96.33 | 100 |
| Time (s) | 10 | 12 | Slightly higher * |
| Space | 6.78 MB | 200 kB < memory < 6.78 MB | Very less |

*Because of additional stage of virtual decompression and block segmentation process.

The proposed page segmentation algorithm works with the assumption that the documents are made noise-free before they are compressed. With the current version of the algorithm, the presence of noise degrades the overall performance of the segmentation. Therefore, to overcome this limitation, a separate pre-processing stage of noise removal in compressed documents is necessary, which could be an interesting extension of this paper. Another strategy to overcome the presence of some small noise, such as salt and pepper noise, is to delete or neglect the black runs of small run-lengths during the segmentation process and to use the concept that merges two longer nearby runs (white runs in the case of vertical column separator detection). Based on this idea, the proposal for conducting a research paper on segmentation in the presence of noise has already submitting to ICDAR2015.

## 5. Conclusion and Scope for Future Work

In this study, a novel method has been proposed to carry out automatic page segmentation in run-length compressed text documents that may have text regions with inverted text colors in part of the document, without having to go through the decompression stage. Our proposed method detects and recovers the inverted text color regions in the compressed documents and performs page segmentation with the following major stages: detecting vertical column separators, computing their related parameters, and segmenting text columns. The text column blocks that are segmented are further decomposed into text blocks, paragraphs, text-lines, words, and characters. Our proposed method has been experimented on and validated with a subset of compressed text documents the from publicly available MARG dataset.

The page segmentation method that we developed is best suited for segmenting compressed documents that contain vertical column separators with the Manhattan layout. Furthermore, our method has been specifically designed to work with text documents. Due to these limitations there is a scope for improving the methods that we have proposed in this paper. The general problem of page segmentation in the presence of noise, skew, complex backgrounds, non-text components, and a non-Manhattan layout could be used as an extension to this paper.